\documentclass{llncs}
\usepackage{makeidx}  % allows for indexgeneration
\usepackage{url}      
\usepackage{subfigure}
\usepackage{amsfonts}
\usepackage{amsmath}
\usepackage{cancel}
\usepackage{graphicx}
\usepackage{color}
\usepackage{hyperref}
\usepackage{tikz}
\usepackage{ifthen}
\usepackage{xcolor}
\usepackage{multirow}
\usetikzlibrary{shadows.blur}

\newlength{\forkmeoffset}
\definecolor{forkmebg}{HTML}{CC0000}
\definecolor{forkmefg}{HTML}{EEEEEE}

\DeclareMathOperator*{\argmax}{arg\,max}

\newcommand{\etal}{\textit{et al}.}

\usepackage{graphicx}
\graphicspath{{pics/}{figs/}}
\begin{document}

\frontmatter          % for the preliminaries
\pagestyle{headings}  % switches on printing of running heads
\mainmatter              % start of your contributions

\title{Straight to the point: reinforcement learning for user guidance in ultrasound}
\titlerunning{Straight to the point: reinforcement learning for user guidance in ultrasound}  % 

\author{
Fausto Milletari,
Vighnesh Birodkar,
Michal Sofka
}

\institute{
  4Catalyzer Corporation
}

\authorrunning{F. Milletari \and \etal}   

\maketitle              % typeset the title of the contribution

\begin{abstract}
Point of care ultrasound (POCUS) consists in the use of ultrasound imaging in critical or emergency situations to support clinical decisions by healthcare professionals and first responders. 
In this setting it is essential to be able to provide means to obtain diagnostic data to potentially inexperienced users who did not receive an extensive medical training. Interpretation and acquisition of ultrasound images is not trivial. First, the user needs to find a suitable sound window which can be used to get a clear image, and then he needs to correctly interpret it to perform a diagnosis. Although many recent approaches focus on developing smart ultrasound devices that add interpretation capabilities to existing systems, our goal in this paper is to present a reinforcement learning (RL) strategy which is capable to guide novice users to the correct sonic window and enable them to obtain clinically relevant pictures of the anatomy of interest. We apply our approach to cardiac images acquired from the parasternal long axis (PLAx) view of the left ventricle of the heart. 

\end{abstract}

\section{Introduction}
\label{sec:intro}
Ultrasound (US) is a flexible, portable, safe and cost effective modality that finds several applications across multiple fields of medicine. 
In particular, ultrasound is widely used to assess the functionality of the heart due to its capability of showing motion in real time allowing clinicians evaluate the overall health of the organ.

The characteristics of ultrasound make it extremely suitable for applications related with emergency medicine and point of care (POC) decision making. 
Recently, several ultra-portable and lightweight ultrasound devices have been announced and commercialized to enable these applications. 
These products have been envisioned to be extremely inexpensive, have a long battery life, a robust design and to be operated by inexperienced users who have not received any formal training. 
In order to reach the latest goal, images need to be interpreted by a computer vision based system and accurate instruction for fine manipulation of the ultrasound probe need to be provided to the user in real time.

Machine learning (ML) and in particular deep learning (DL) have been recently employed to solve a diverse set of problem in medical image analysis. 
The performances of image segmentation, interpretation, registration and classification algorithms were significantly boosted by the usage of deep learning based techniques which reached or even surpassed the accuracy of human raters. 

In this paper we show how to use deep learning and in particular deep reinforcement learning to create a system to guide inexperienced users towards the acquisition of clinically relevant images of the heart in ultrasound. 
We focus on acquisition through the parasternal long axis (PLAx) sonic window on the heart which is one of the most used views in emergency settings due to its accessibility. 

In our acquisition assistance framework the user is asked to place the probe anywhere on the left side of the patient's chest and receives instructions on how to manipulate the probe in order to obtain a clinically acceptable parasternal long axis scans of the heart. Every time an image is produced by the ultrasound equipment, our deep reinforcement learning model predicts a motion instruction that is promptly displayed to the user. In this sense, we are learning a control policy that predicts actions (also called instructions) in correspondence of observations, which makes reinforcement learning a particularly suitable solution.
This problem has several degrees of freedom. Apart from instructions regarding left-right and top-bottom motions, the user will also receive fine-grained manipulation indications regarding rotation and tilt of the probe.

Our reinforcement learning approach is trained end to end, meaning that differently from other techniques we train our model to solve the acquisition assistance task directly. For this reason we need to capture exemplary training data in a way that allows us to simulate acquisition assistance offline, since it would be impractical to do this in real time on real people while training. We build environments to simulate offline any (reasonable) acquisition trajectory the user may take while scanning the patient by making use of tracked video sequences.
The paradigm we choose for deep RL is DQN coupled with an epsilon greedy exploration strategy and delayed updates. In \cite{mnih2013playing} a similar approach, making use of a simpler network architecture, has shown excellent performances on arcade video-games.

We compare the performances of our method with the ones obtained by training a classifier to learn a policy on the same data in a fully supervised manner. 

\section{Related Work}
\label{sec:relatedwork}
Reinforcement learning has been recently employed to solve several computer vision related problems and specifically to achieve superhuman performances in playing ATARI games and 3D video-games such as "Doom" \cite{lample2017playing}. 

In \cite{mnih2013playing,mnih2015human} a convolutional deep neural network has been employed together with Q-learning to predict the expected cumulative reward $Q(s,a)$ associated with each action that the agent can perform in the game. 
In \cite{lillicrap2015continuous} a learning strategy that employs two identical networks, updated at different paces, is presented. In this paper, the target network is used for predictions and is updated smoothly at regular intervals, while the main network gets updated batch-wise through back-propagation. This is particularly useful in continuous control. 
In \cite{wang2015dueling} the network architecture used to predict the Q-values is modified to comprise two different paths which predict, respectively, the value $V(s)$ of being in a certain state and the advantage of taking a certain action in correspondence to that state. This strategy has resulted in increased performances. 
In \cite{van2016deep} target Q-values, which are learned during training, are computed differently than in \cite{mnih2013playing}. Instead of having the network regress Q-values computed as the reward $r_t$ plus $\gamma\argmax_a{Q^{*}(s_{t+1},a)}$, they use $r_t + \gamma Q^{*}(s_{t+1},a_{t+1})$. The main difference is that, in the latter, the action $a_{t+1}$ is the one that is selected by the network in correspondence of the state $s_{t+1}$, and not $a$ which is the one yielding the maximum Q-value. This yields increased stability of the Q-values. 

Reinforcement learning has been applied in medical domain for the first time in \cite{ghesu2016artificial} to solve a landmark localization problem in MRI with a game-like strategy as the problem may resemble a maze navigation problem. In \cite{neumann2016self} a similar approach has been applied to heart model personalization on synthetic data.

In this work we apply reinforcement learning to a guidance problem whose goal is to provide instructions to users in order to enable them to scan the left ventricle of the heart using ultrasound through the parasternal long axis sonic window. We build our learning strategy to perform end-to-end optimization of the guidance performances. In this work we show the details of our learning environment, which allows the agent to obtain simulated ultrasound acquisitions and we compare our system with supervised policy learning via classification.

\section{Method}
\label{sec:method}
\begin{figure} 	
\centering 	
\includegraphics[scale=0.35]{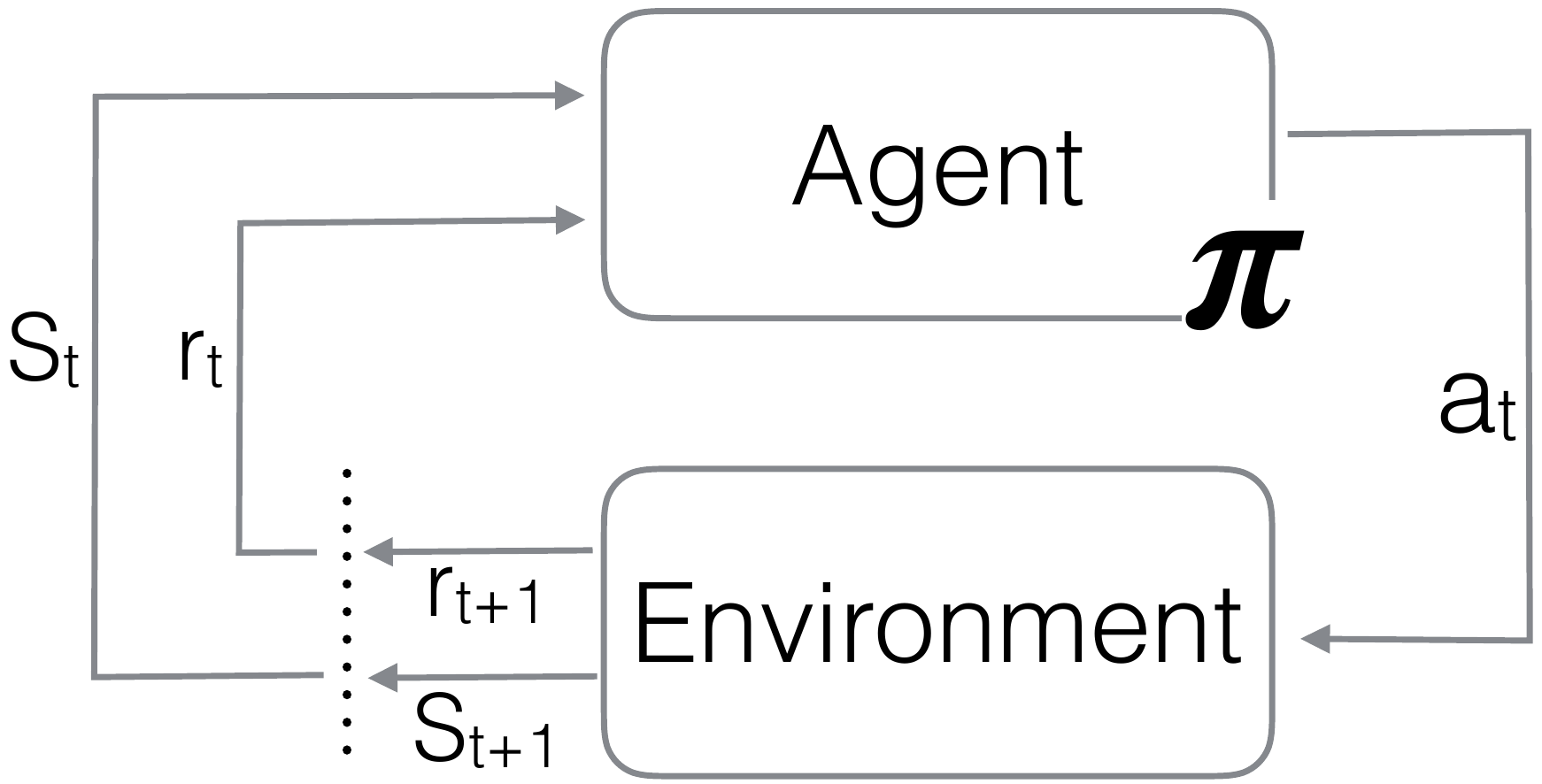} 	
\caption{Schematic representation of the reinforcement learning framework.} \label{fig:MDP} 
\end{figure}

As previously discussed, reinforcement learning models are usually trained by letting an agent interact with the environment (Eg. play a game) using the current policy in order to obtain new observations and rewards. Once new experiences are obtained, the policy can be updated by learning from them.
More formally the reinforcement learning problem is usually formulated as a Markov decision process (MDP) (Figure \ref{fig:MDP}). At each point in time, the agent observes a state $S_t$ and interacts with the environment, using its policy  $\pi \in \Pi$, through actions $a \in A$ obtaining a finite reward $r_t$ together with a new state  $S_{t+1}$. $\Pi$ is the set of all possible policies while $A$ is the set of all supported actions.

The set of supported actions, in our system, contains $9$ actions as shown in table \ref{tab:actions}.

\begin{table}
\caption{Set of actions supported by the agent. These action are mapped to the corresponding effect in the simulated acquisition framework.}

\label{tab:actions}
\centering{}%
\begin{tabular}{|c|c|}
\hline 
Action & Effect\tabularnewline
\hline 
\hline 
NOP & Stops the virtual probe. Should be issued at correct view. \tabularnewline
\hline 
Move Lateral & Translates the probe towards the patient's left\tabularnewline
\hline 
Move Medial & Translates the probe towards the patient's right\tabularnewline
\hline 
Move Superior & Translates the probe towards the patient's head \tabularnewline
\hline 
Move Inferior & Translates the probe towards the patient's feet\tabularnewline
\hline 
Tilt Supero-laterally & Tilts the probe towards the head of the patient\tabularnewline
\hline 
Tilt Infero-medially & Tilts the probe towards the feet of the patient\tabularnewline
\hline 
Rotate Clockwise & Rotates the probe clockwise\tabularnewline
\hline 
Rotate Counter-Clockwise & Rotates the probe counter-clockwise\tabularnewline
\hline 
\end{tabular}
\end{table}

In this section we present the details of our implementation. First we discuss the implementation of the acquisition simulation environment which is necessary for learning a policy via reinforcement learning. Then we give the details of our DQN implementation and of the convolutional architecture we employ in this work. Last we  introduce the fully supervised strategy which is used in this work to obtain means of comparison. 

\subsection{Simulated acquisition environment}
In order to learn from experience, our reinforcement learning agent needs to collect data according to its policy by physically moving the probe on the chest of the patient in order to obtain data and rewards. It is unfortunately impossible to implement such a system in practice due to the fact that acquiring the trajectories would take an enormous amount of time and a person would need to be scanned for the whole duration of learning. 

\begin{figure} 	
\centering 	
\includegraphics[scale=0.35]{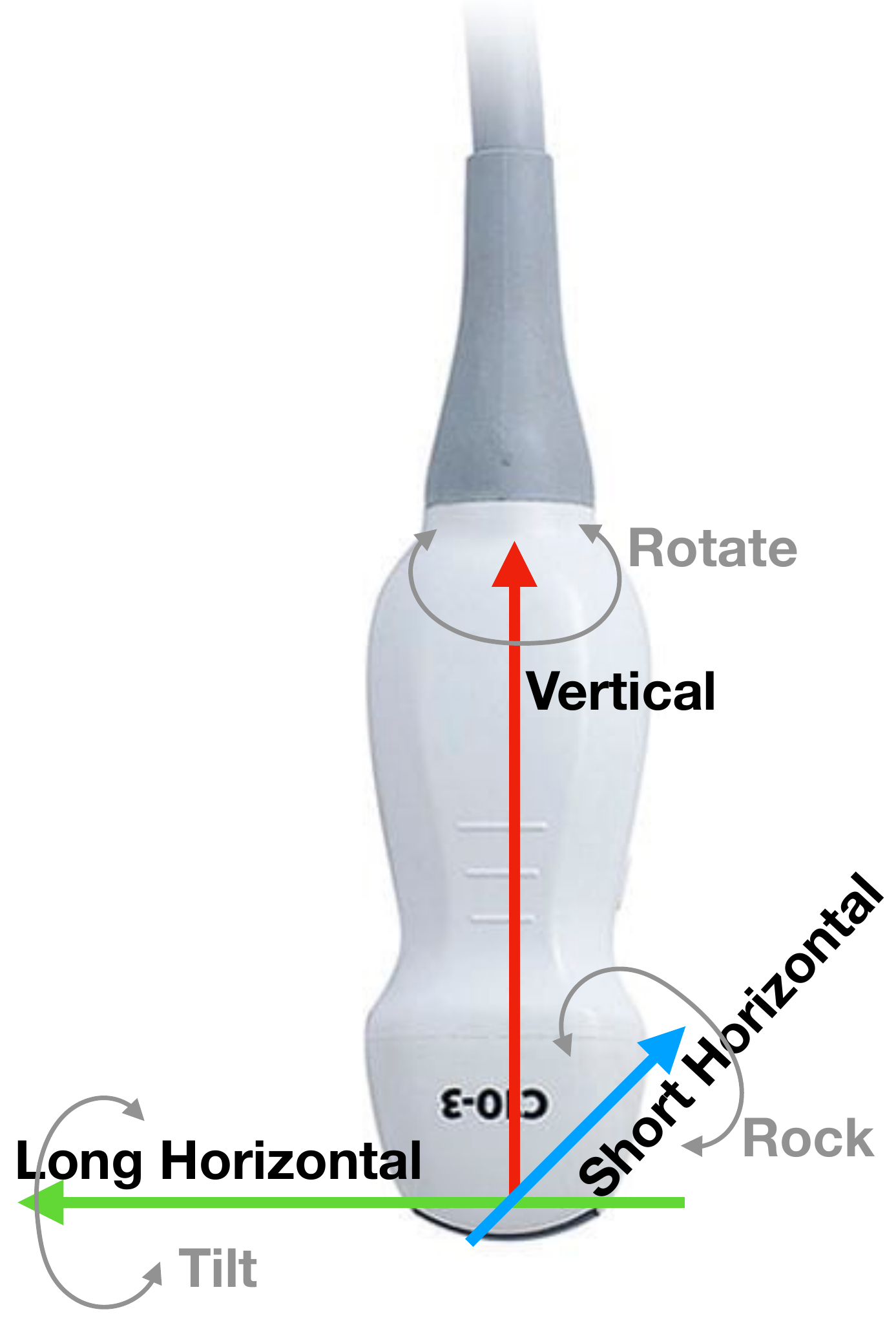} 	
\caption{Degrees of freedom of the probe during acquisition. We cover all degrees of freedom apart from rotations about the short horizontal axis (rocking).} \label{fig:DOF} 
\end{figure}

We have resorted to acquiring, independently from our learning procedure, a large number of spatially tracked video frames from patients. By drawing spatial relationships between the frames, we are able to navigate the chest area offline and obtain simulated trajectories. We have defined, for each participant in the study, a work area covering a large portion of their chest. We have divided this area into $7\times7$ mm spatial bins. The bins from which it is possible to obtain a valid PLAx view by fine manipulation of the probe, are annotated as "correct" while all other bins remain unmarked. This annotation is necessary to implement the reward system.

Our system offers guidance for 4 out of the 5 degrees of freedom of probe motion (Figure \ref{fig:DOF}). We get data for the first two degrees of freedom, left-right and top-bottom translations, by moving the probe in a regular and dense grid pattern over the chest in order to "fill" each bin of the grid with at least $25$ frames. In correspondence of the bins marked "correct", the sonographer is also asked to acquire $50$ "correct" frames, showing the best view and $50$ frames from each of the following scenarios: the probe is rotated by an excessive amount in the (i) clockwise or (ii) counterclockwise direction, or the probe is tilted by an excessive amount in the (iii) infero-medial or (iv) supero-lateral direction. In this way data for the last two degrees of freedom is obtained.

When the agent interacts with the environment it is free to move in any direction and reach any bin within the grid. For the bins that are marked "correct" it is additionally able to apply rotations and tilts. This limitation is only introduced because of the fact that acquiring all rotations and tilts for all grid bins would be too time consuming to be done in practice. Moreover, such fine manipulation instructions are meaningful only when the user places the probe near one of the "correct" views.

\begin{figure} 	
\centering 	
\includegraphics[scale=0.35]{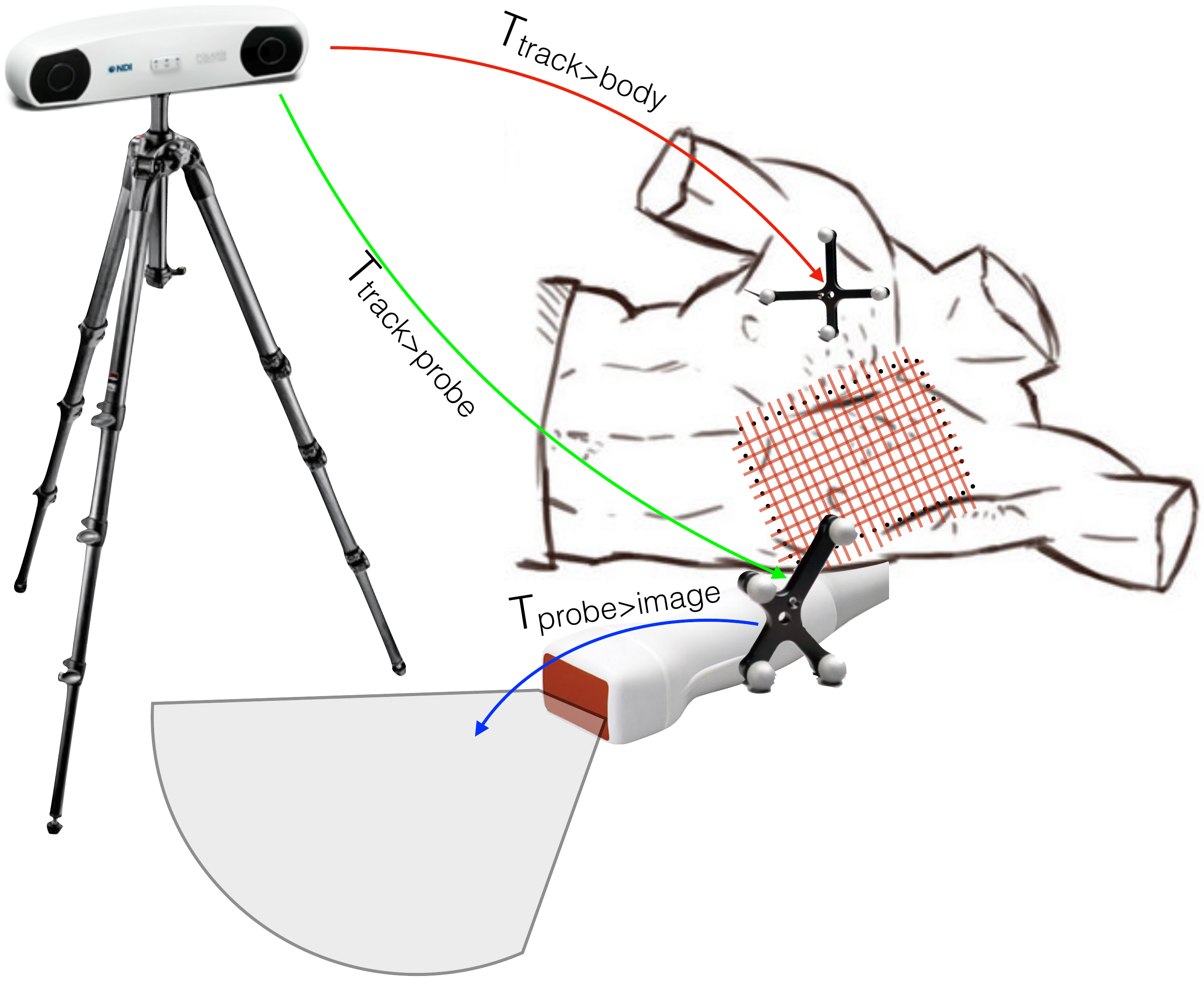} 	
\caption{Schematic representation of our data acquisition system which comprises a probe and a tracking system in order to obtain tracked video frames from the patient.} \label{fig:tracking} 
\end{figure}

In order to build the environment we need to track both the body of the patient and the probe as data gets acquired. A schematic representation of our tracking system is shown in Figure \ref{fig:tracking}. The tracking system, a NDI Polaris Vicra optical tracker, produces in real time a tracking stream consisting of two $4\times4$ transformation matrices $T_{track>probe}$ and $T_{track>body}$. The transform $T_{probe>image}$, which is necessary to obtain the true pose of each picture acquired through our system, is obtained by performing calibration with the open source software fCal, which is provided as part of the PLUS framework \cite{lasso2014plus}. 
The video frames are acquired through an ultrasound probe and supplied to the data acquisition system through OpenIGTlink interface \cite{tokuda2009openigtlink}. 
The tracking and video streams are handled and synchronized using the PLUS framework in order to obtain tracked frames.

Each bin of the environment contains the necessary data, it is possible to interact with it by performing actions which result in state changes and rewards. The actions can have the effect of either stopping the virtual probe ("NOP" action), bringing it closer or further away from the nearest goal point.

At the beginning of each episode the environment is reset and a virtual "probe" is randomly placed in one of the bins. When the agent request an action that is incompatible with the structure of the grid, for example by requesting grid bins that are outside the work area or request tilts and rotations in correspondence of bins that have not been marked "correct", the environment supplies a random frame from the current bin and the same reward that would have been given in correspondence of an action that would have brought the virtual probe further from the nearest goal point. 

A summary of the agent reward scheme used in this paper is provided in Table \ref{tab:rewards}.

\begin{table}
\caption{Reward strategy for our reinforcement learning agent.}
\begin{centering}
\begin{tabular}{|c|c|c|}
\hline 
 & Correct Bin & Other Bin\tabularnewline
\hline 
Stop & $1$ & $-0.25$\tabularnewline
\hline 
Closer & --- & $0.05$\tabularnewline
\hline 
Further & $-0.10$ & $-0.10$\tabularnewline
\hline 
\end{tabular}
\par\end{centering}
\label{tab:rewards}
\end{table}

\subsection{Deep Q-Network}
In this work we implement the Q-learning paradigm already employed by \cite{mnih2013playing,mnih2015human}. This off-policy learning strategy leverages a convolutional neural network to regress Q-values which are the expected cumulative rewards associated with each action in correspondence of a state. As previously stated, the input of the model are ultrasound images, and its output is represented by nine Q-values, one for each action. 
Similarly to \cite{lillicrap2015continuous} we instantiate two copies of the same network. We have a target network which produces the values $Q_{\theta^{*}}(s,a)$ and a main network which predicts $Q_{\theta}(s,a)$. 

\begin{figure} 	
\centering 	
\includegraphics[scale=0.50]{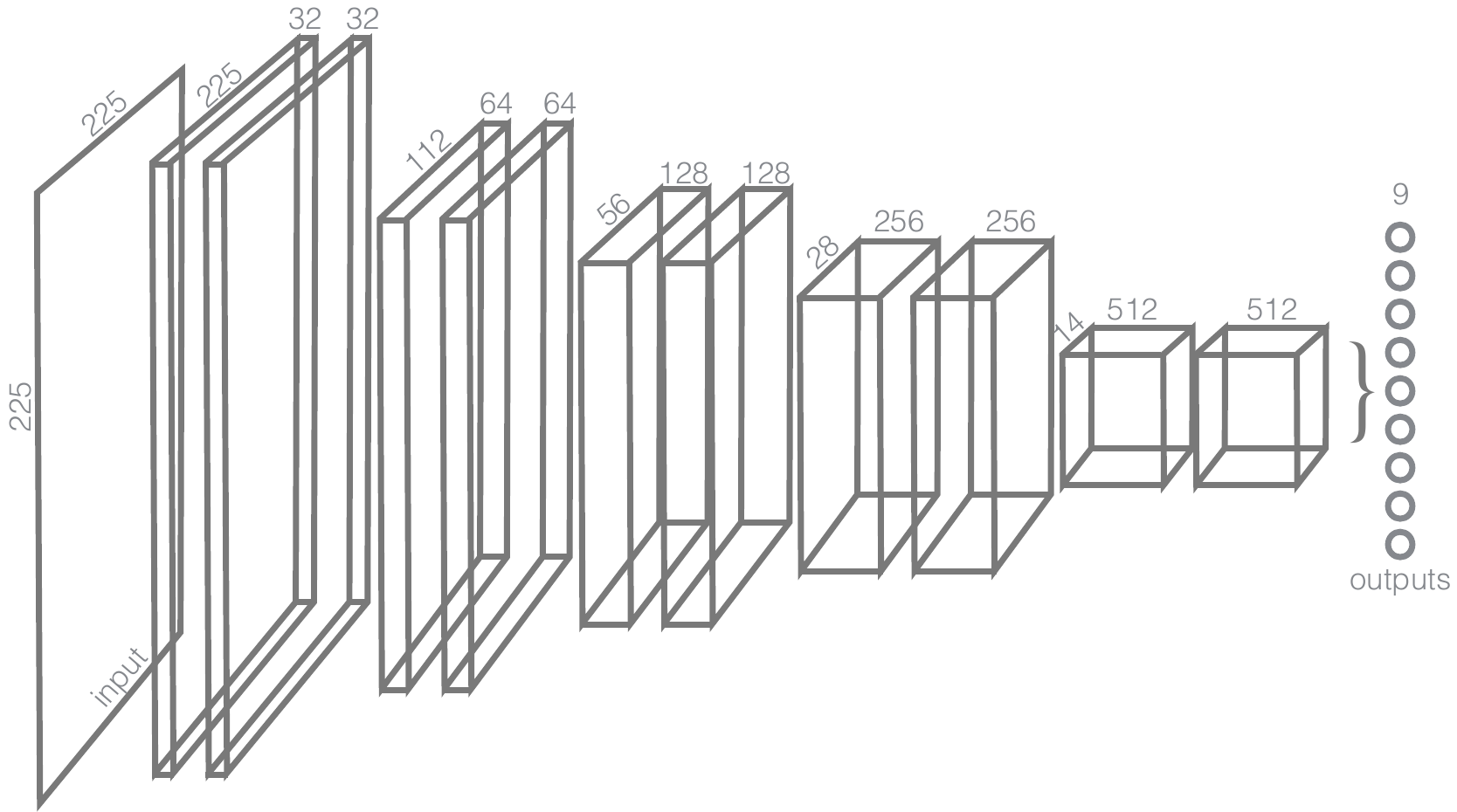} 	
\caption{Schematic representation of the network architecture.} \label{fig:architecture} 
\end{figure}

In order to train our agent we interact with the training environments. Each environment refers and represents to one patient. During an episode, we select an environment among those available for training and we reset the virtual probe to a random position. We then use the main network to collect experience by interacting with the environment. We implement exploration using an epsilon-greedy strategy which randomly hijacks and replaces the actions chosen through $\argmax_a(Q_{\theta}(s,a))$ with random ones. In this way we are able to balance the needs for exploring the environment and the follow the learned policy. All agent's experiences are collected in an experience replay buffer of adequate size as previously done in \cite{mnih2013playing}. Since all our data is pre-acquired it is possible to increase the memory efficiency of the experience replace buffer by storing in memory image paths on the file system instead of storing uncompressed images. 

Once there is enough data in the experience replay buffer, we sample random training batches from it and we use them to update the parameters $\theta$ of the main network using back-propagation. The objective function that we minimize with respect to the parameters of the network, using ADAM as our optimizer, is

\[
C(\theta,s_{t},a_{t})=\left\Vert Q_{\theta}(s_{t},a_{t})-T(s_{t},a_{t})\right\Vert _{2}^{2}
\]
\[
T(s_{t},a_{t})=r_{t}+\gamma\argmax_a(Q_{\theta^{*}}(s_{t+1},a))
\]

The target network network is trained with a different strategy: the parameters $\theta^{*}$ of the target network are updated with the parameters of the main network once every $250$ episodes.

A schematic representation of the network architecture is shown in Figure \ref{fig:architecture}. This network makes use of global average pooling \cite{lin2013network} applied after the output of the last convolutional layer. All the non-linearities employed throughout the network are exponential linear units (ELU) \cite{clevert2015fast}. A 9-way fully connected layer follows the global average pooling and produces the outputs of the network. 

During testing, the target network interacts with the environment. All actions are chosen deterministically through $\argmax_a(Q_{\theta^{*}}(s,a))$ which is, therefore, a stationary deterministic policy.

\subsection{Supervised policy learning}

In order to obtain means of comparison for our approach we have implemented a supervised policy learning approach which relies on classification and labeled data to learn the right action to perform in correspondence of each state. 
When we acquire data from patients we build environments where the parameters of the correct view in terms of translation, rotation and tilt are known. This enables us to label each image in each bin of the grid with one action, which would be the optimal action to perform in that state if we rely only on the Manhattan distance $abs(\mathbf{x} - \mathbf{x}_{goal})$ between the bin position $\mathbf{x}$ on the grid and the goal bin position $\mathbf{x}_{goal}$. In particular, for each bin of the grid, we choose the label for its images as the action that reduces the distance to the goal on the axis where the distance is currently the smallest. 

We train a classifier with the same architecture shown in Figure \ref{fig:architecture}, with the only exception that the last layer is followed by a soft-max activation function. We use all the data that is available to our reinforcement learning agent, shuffled and organized in batches of the same size of the ones used for our DQN. 

During testing we use the same environments used by the reinforcement learning agent to test the supervised policy end-to-end on the guidance task. In this way we can compare on fair grounds the performances of the two strategies.

\section{Results}
\label{sec:results}
We evaluate our method on the end-to-end guidance task described in the previous sections, using one environment for each patient. We train our approach on $22$ different environments corresponding to circa $160$ thousand ultrasound images, and we test our approach on $5$ different environments which contain circa $40$ thousand scans. During testing with start from each and every grid bin of each environment and we test the guidance performances of the approach.

As previously explained, we train both a RL-based approach and a supervised classification-based approach. Results are shown in Table \ref{tab:rewards}.

We perform data augmentation for both the supervised and RL approaches. Each training sample is slightly rotated, shifted and re-scaled by a random quantity before being presented as input to the network. Also the gamma of the images is subject to augmentation. 
The episodes have a standard duration of 50 steps and "NOP" operations do not terminate the episode. Instead, a new, randomly selected, image from the same grid bin is returned to the agent. 

\begin{table}
\caption{Summary of performance of RL approach versus supervised approach on the test data-set.}
\begin{centering}
\begin{tabular}{|c|c|c|}
\hline 
 & {Reinforcement Learning} & \multirow{2}{*}{Supervised}\tabularnewline
\cline{1-2} 
Performances & Continuous & \tabularnewline
\hline 
\hline 
Correct guidance & \textbf{86.1\%} &  77.8\%\tabularnewline
\hline 
Incorrect guidance & \textbf{13.9\%} &  22.2\%\tabularnewline
\hline
Incorrect NOP percentage & \textbf{1.6\%} &  25.9\%\tabularnewline
\hline 
\hline 
Behaviour & \multicolumn{2}{c|}{}\tabularnewline
\hline 
\hline 
Avg. number negative rewards & \textbf{30.3}\% &  36.9\%\tabularnewline
\hline 
Avg. number positive rewards & \textbf{69.6\%} &  63.1\%\tabularnewline
\hline 
\end{tabular}
\par\end{centering}
\label{tab:results}
\end{table}

Our results are summarized in Table \ref{tab:results}. The table is split in two parts: the first part summarizes the performances of the method on the end-to-end guidance task and inform us on the percentage of correct and incorrect guidance. That is, the percentage of episodes that have ended in a "correct" bin. Additionally we report the percentage of "NOPs" that have been issued at an incorrect location. Please note that "NOP" can be issued multiple times during a single episode. The agent may have briefly issued an "incorrect NOP" even during successful episodes. The evaluation reveals that the supervised approach is less successful than the RL approach on the guidance task.
The second part of the table reveals information about the behaviour of the reward. Also these results demonstrate that our RL agent is performing more "correct" actions than its supervised counterpart.

\section{Conclusion}
We have developed a framework to interpret ultrasound images with the objective of guiding an inexperienced user to acquire adequate images of the heart from the PLAx sonic window. Our approach has proven to achieve better results than a supervised approach trained on the same data and tested on the end-to-end guidance task. The intuition behind this is that RL is able to avoid and go around areas that are highly ambiguous as the Q-Values in correspondence of the actions leading to those states should not be very high. Moreover RL implicitly learns the spatial arrangement of the different pictures on the chest. The Q-values of grid bins that are very far from the goal are much lower than the Q-values obtained for states nearby the correct sound window. This means that, although the best actions for these bins may be the same, the network must be able to distinguish them since in order to produce different Q-values ranges. 

Although the results have shown to be promising there are still issues related with this method. Scalability of data acquisition strategy is the first issue. In order to acquire training data it's necessary to undergo a tedious scanning procedure that can last up to 20 minutes per patient. Moreover, it requires dedicated and costly tracking equipment which work only when the line of sigh (LoS) between the tracker and the markers is not occluded. The resulting environments allow acquisition trajectory simulation but, of course, the images they provide slightly differ from the ones obtained by a true, live scan.

The training time required by the presented reinforcement learning approach is also very long, and limits the possibility of experimenting with different sets of hyper-parameters. 

In conclusion, we believe that this method is the one of the first step to converge towards a solution which aims to solve the guidance task end-to-end in a more reliable and effective manner.

\bibliographystyle{splncs03}
\bibliography{bibliography}

\end{document}